# EMBRACE: A Multi-task Framework for Comprehensive Quality Assessment in Cleavage-stage Embryo

Anwar Hussain Sofi[a], Jung-Hua Wang[b,c*], Ming-Jer Chen[d], Tsung-Hsien Lee[e], Yu-Chiao Yi[f], Ming-Kuan Lin[c], Yi-Chung Lai[b]

[a] International Master Program in Applied Artificial Intelligence, National Taiwan Ocean University, Keelung, 202301, Taiwan

[b] AI Research Center, National Taiwan Ocean University, Keelung 20224, Taiwan

[c] Department of Electrical Engineering, National Taiwan Ocean University, Keelung 20224, Taiwan (e-mail: jacksonlin092@gmail.com).

[d] Department of Obstetrics and Gynecology, Lee Women's Hospital, Taichung 40652, Taiwan

[e] Department of Obstetrics and Gynecology, Chung Shan Medical University Hospital, Taichung 40201, Taiwan

[f] Department of Obstetrics and Gynecology, Taichung Veterans General Hospital, Taichung 407219, Taiwan

[*]Corresponding author: jhwang@email.ntou.edu.tw, ORCID ID:0000-0003-1769-7396, Keelung, 202301, Taiwan

## Abstract

Cleavage-stage embryo assessment in in vitro fertilization requires the integrated interpretation of cytoplasmic fragmentation, developmental stage, and blastomere symmetry. However, conventional visual assessment is affected by observer variability, particularly when fragmented regions are small, irregular, or low contrast. This study presents EMBRACE, a multi-task deep learning framework for jointly performing cytoplasmic-fragmentation segmentation, t2/t4 developmental-stage classification, and blastomere-symmetry grading from static cleavage-stage embryo microscopy images. EMBRACE combines a shared ResNet-50 backbone, a concatenation-based multi-scale feature-fusion (C-MSFF) module, a U-Net-style segmentation decoder, and two task-specific classification heads. After predefined inclusion and exclusion criteria, 9,137 annotated embryo images were divided into 7,309 training, 914 validation, and 914 held-out test images. On the held-out test set, EMBRACE achieved a Dice coefficient of 0.781 and an intersection over union of 0.677 for fragmentation segmentation. Developmental-stage classification achieved an accuracy of 0.995, macro-F1 of 0.994, and AUC of 1.000. Blastomere-symmetry grading achieved a balanced accuracy of 0.901, macro-F1 of 0.907, and quadratic weighted kappa of 0.859. These findings support the feasibility of combining spatially inspectable fragmentation localization with embryo-level morphology assessment in a single framework. External and prospective validation is required before clinical deployment.

## 1. Introduction

Embryo assessment remains a central decision point in In Vitro Fertilization (IVF), where laboratory evaluation supports decisions related to embryo transfer, continued culture, and cryopreservation. In routine embryology practice, embryo prioritization commonly relies on non-invasive morphological features, including blastomere number, cytoplasmic appearance, fragmentation, blastomere regularity, and developmental progression [1,2]. Although embryo developmental competence is influenced by biological, laboratory, and patient-specific factors, morphology-based assessment remains clinically important because it provides immediate visual information without additional embryo manipulation.

At the cleavage stage, embryo evaluation requires the combined interpretation of multiple morphology-related features rather than reliance on a single visual cue. Developmental cell stage reflects cleavage progression and provides temporal context for embryo assessment. Cytoplasmic fragmentation represents anucleate cytoplasmic material and has been associated with reduced implantation and pregnancy potential. Blastomere symmetry reflects the regularity of early cell division, and unevenly sized blastomeres have been associated with lower implantation and pregnancy rates [4]. Evidence linking Day 2 blastomere symmetry with later blastocyst grade and ploidy status further supports the relevance of symmetry assessment during early embryo development [5]. Early cleavage timing is also associated with later blastocyst formation and embryo quality [6], while cleavage anomalies have been linked with embryo survival after prolonged culture [7]. Together, these findings support the continued relevance of integrated cleavage-stage morphology assessment.

Morphokinetic literature [8] indicates that early developmental kinetics, specifically tPNf, t2, and t4, correlate significantly with embryogenesis, offering critical temporal indices for embryo viability assessment. However, because pronuclear membrane breakdown (tPNf) precedes the manifestation of distinct two- or four-cell blastomere topologies, it is incompatible with the proposed framework, which necessitates visible blastomeric structures for symmetry grading and localized cytoplasmic fragments for segmentation. Conversely, isolating t2 and t4 is clinically and biologically justified, as these specific cleavage milestones establish the essential temporal context required to evaluate concurrent fragmentation and blastomere organization. Consequently, a multi-task computational framework capable of concurrently analyzing fragmentation, blastomere symmetry, and t2/t4 developmental stages aligns more robustly with the multi-dimensional nature of cleavage-stage embryo grading than conventional single-output prognostic models.

Despite its clinical utility, conventional visual morphology assessment remains affected by subjectivity and observer variability. Inter- and intra-observer differences have been reported in early embryo morphology assessment [9], and variability has also been documented in time-lapse annotation [10]. Fragmentation assessment is particularly challenging because fragmented regions may be small, scattered, low-contrast, irregular, or visually similar to surrounding cytoplasmic texture. Subtle blastomere size inequality can also be difficult to judge consistently in borderline embryos. These challenges motivate objective and visually inspectable computational tools that can support embryo image review without replacing expert judgment.

Artificial intelligence (AI) and deep learning have increasingly been explored for embryo and oocyte classification [11], static-image embryo viability prediction, and time-lapse-based blastocyst formation and quality prediction. Static microscopy-based AI models have demonstrated the feasibility of predicting embryo viability from optical images [12], while time-lapse deep learning approaches have been used to predict blastocyst formation and embryo quality from developmental image sequences [13]. However, many existing AI approaches remain centered on image-level classification or single-endpoint prediction. For fragmentation assessment, this is a critical limitation because a probability score or embryo-level grade alone cannot show whether the model has localized the correct fragmented regions.

Segmentation-based analysis addresses this interpretability gap by offering pixel-level masks that can be visually inspected and compared with embryo images, as demonstrated in interpretable embryo-analysis pipelines [14]. Biomedical segmentation architectures provide an established foundation for pixel-level localization through encoder–decoder designs with spatial recovery pathways [15]. In embryo fragmentation assessment, spatial localization is important because fragmented regions may appear as small anucleate structures distributed between blastomeres or visually overlap with cytoplasmic texture. Dice-based objectives are useful for sparse foreground segmentation [16], while IoU-oriented objectives directly support region-overlap optimization [17]. Boundary-aware objectives are relevant when fragment boundaries are visually uncertain or highly imbalanced [18]. Therefore, segmentation is not only a technical output but also an interpretability mechanism that can make fragmentation-related predictions more transparent to embryologists.

Many embryo AI pipelines still focus on separate image-level prediction tasks, such as embryo/oocyte classification, static-image embryo viability prediction, or time-lapse blastocyst prediction. Fragmentation segmentation requires local and boundary-sensitive analysis, whereas developmental-stage classification and blastomere symmetry grading require global interpretation of blastomere number, size, and spatial arrangement. If these outputs are predicted through separate models, the resulting pipeline may be less efficient and may not learn a shared embryo representation. Multi-task learning provides a structured alternative by allowing related tasks to share visual representations [19] while preserving task-specific prediction heads. However, task compatibility must be evaluated carefully when dense segmentation and image-level classification impose different optimization demands on a shared model [20]. Balancing loss functions can also be controlled by hyperparameters to affect multi-task optimization [22]. Gradient conflict [21] may further reduce performance when competing tasks update the shared encoder in different directions.

These considerations motivate EMBRACE, a multi-task deep learning framework for cleavage-stage embryo morphology assessment. EMBRACE jointly predicts a pixel-wise fragmentation mask, a t2/t4 developmental-stage label, and a blastomere symmetry grade. The framework combines shared representation learning with task-specific outputs so that local fragmentation evidence and global embryo morphology can be analyzed within one model. This design is intended to better reflect the integrated nature of cleavage-stage embryo assessment, where embryologists consider fragmentation, developmental stage, and blastomere symmetry together rather than as isolated findings.

The main objective of this study is to develop and evaluate a compact, spatially interpretable, and clinically aligned multi-output framework for cleavage-stage embryo image analysis. EMBRACE uses a shared feature encoder [23] and Multi-Scale fusion representation [24, 38] to support fragmentation segmentation, t2/t4 stage classification, and blastomere symmetry grading. The model is optimized using imbalance-aware [25] and boundary-sensitive loss components to address sparse fragmentation foregrounds and class imbalance across morphology labels. By integrating pixel-level fragmentation localization with embryo-level morphology prediction, this study aims to reduce the gap between isolated AI prediction tasks and the multi-feature reasoning used in embryology practice. The framework is intended as a decision-support approach for research and embryo image review, and further external, prospective validation remains necessary before clinical deployment.

## 2. Related work

Conventional embryo morphology assessment is grounded in visible developmental features, including blastomere number, cytoplasmic appearance, cytoplasmic fragmentation, blastomere regularity, and developmental progression. These features are clinically practical because they can be assessed non-invasively using routine microscopy, but they remain imperfect indicators of developmental potential and should not be interpreted as deterministic predictors of implantation, pregnancy, or live birth. Fragmentation and blastomere irregularity are particularly important in cleavage-stage assessment because fragmentation has been associated with reduced implantation and pregnancy potential, while uneven blastomere size has been associated with lower pregnancy and implantation rates. Evidence linking Day 2 blastomere symmetry with later blastocyst grade and ploidy status further supports the relevance of symmetry assessment during early embryo development [5].

Artificial intelligence and deep learning have increasingly been investigated in reproductive medicine for embryo and oocyte classification, embryo viability prediction from static microscopy images, and blastocyst formation or quality prediction from time-lapse data. Earlier AI studies showed that embryo and oocyte images contain quantifiable visual information that can be modeled computationally. More recent studies have demonstrated the feasibility of deep learning for predicting embryo viability from static optical microscopy images and for predicting blastocyst formation and quality from time-lapse monitoring. However, many embryo AI systems remain centered on image-level outputs, which limits their ability to explain the spatial visual evidence supporting a prediction. This limitation is particularly relevant for fragmentation assessment, where the location, extent, and boundary of fragmented regions are clinically meaningful and cannot be fully represented by a single image-level score.

Segmentation-based analysis provides a more spatially interpretable approach by producing pixel-level masks that can be visually compared with the embryo image and expert annotation. U-Net-style encoder–decoder architectures provide a methodological foundation for biomedical image segmentation because they combine contextual feature extraction with spatial recovery through skip connections [15]. In embryo fragmentation assessment, segmentation is challenging because fragmented regions may occupy a small proportion of the image and may have ambiguous boundaries against cytoplasmic texture. Therefore, pixel accuracy is not sufficient as a primary metric for this task because high background dominance can mask poor foreground segmentation performance. Dice-based overlap measures are appropriate for sparse segmentation targets [16], while IoU-oriented objectives support region-overlap optimization [17]. Boundary loss is useful for uncertain boundaries in highly imbalanced segmentation, and focal loss helps emphasize hard examples in dense prediction tasks [26].

Multi-task learning is relevant to cleavage-stage embryo assessment because the clinical task involves both local and global morphology interpretation. Fragmentation segmentation requires local and boundary-sensitive evidence, whereas developmental-stage classification and blastomere symmetry grading require global interpretation of blastomere number, size, and spatial arrangement. Shared representation learning can improve efficiency and promote consistency across related outputs. However, task compatibility must be evaluated rather than assumed [20], because gradient conflict and loss weighting [22] can influence whether joint learning improves or harms individual tasks. EMBRACE addresses this gap by combining pixel-level fragmentation segmentation with embryo-level t2/t4 stage classification and blastomere symmetry grading in one shared-encoder framework, while evaluating whether added architectural complexity improves or harms the balance among segmentation and classification tasks.

The main research gap addressed in this study is the mismatch between the multi-feature nature of cleavage-stage embryo assessment and the image-level design of many embryo AI pipelines. Existing studies have focused on embryo/oocyte classification, static-image viability prediction, time-lapse blastocyst prediction, or isolated interpretable morphology outputs. Clinically, cleavage-stage assessment requires combined interpretation of cytoplasmic fragmentation, developmental cell stage, and blastomere symmetry. However, image-level AI outputs may not provide spatially inspectable evidence for fragmentation, while standalone segmentation may not incorporate developmental-stage or symmetry context. This creates a methodological gap that motivates a shared multi-task framework linking pixel-level fragmentation localization with embryo-level morphology prediction. This study makes the following contributions:

- **Unified multi-task embryo assessment:** EMBRACE is a unified multi-task deep learning framework for simultaneous fragmentation segmentation, developmental-stage classification, and blastomere symmetry grading from static cleavage-stage embryo microscopy images.

• **Spatially inspectable fragmentation assessment:** EMBRACE integrates pixel-level fragmentation localization with embryo-level morphology prediction, allowing fragmentation-related outputs to be visually inspected rather than inferred only from global image-level scores.

• **Shared multi-scale representation learning:** The proposed framework utilizes a shared encoder coupled with a Concatenation-Based Multi-Scale feature fusion representation to effectively integrate localized, boundary-sensitive fragmentation features with high-level semantic embryo morphology cues. Although ResNet-50 is employed throughout this study as the primary feature extraction backbone for demonstrative purposes, the architecture is inherently modular and backbone-agnostic, permitting the seamless substitution of alternative image classification networks depending on computational or performance requirements.

• **Imbalance-aware and boundary-sensitive optimization:** The segmentation branch applies a compound loss formulation combining weighted binary cross-entropy, Dice loss, Lovász loss, boundary loss, and focal loss to address sparse fragmentation foregrounds, overlap quality, IoU-oriented optimization, uncertain fragment boundaries, and hard examples.

• **Large annotated embryo dataset:** A large-scale manual annotation effort was performed using LabelMe on more than 10,000 cleavage-stage embryo images, including pixel-level cytoplasmic fragmentation annotations. After predefined inclusion and exclusion criteria, 9,137 eligible annotated t2/t4 embryo images were used for supervised model development and evaluation.

• **Systematic ablation across model variants:** We compare EXP1-EXP6 model variants under the same dataset split and evaluation protocol to examine whether added architectural complexity improves or harms the balance among fragmentation segmentation, developmental-stage classification, and blastomere symmetry grading. This is important because multi-task learning can benefit from shared representations, but task compatibility must be evaluated carefully, and gradient conflict may reduce performance when tasks compete during shared optimization.

• **Comprehensive model evaluation:** We evaluate model behavior using segmentation metrics, classification metrics, agreement statistics, confusion matrices, qualitative overlays, ablation analysis, and failure-case review to provide a cautious and reproducible assessment of the proposed framework.

## 3. Methodology

### 3.1 Dataset preparation

The proposed EMBRACE pipeline processes a single static cleavage-stage embryo microscopy image and generates three coordinated outputs: a pixel-wise cytoplasmic fragmentation mask, a t2/t4 developmental-stage prediction, and a blastomere symmetry grade. This multi-output design reflects the clinical structure of cleavage-stage morphology assessment, where fragmentation, developmental progression, and blastomere symmetry are interpreted together rather than as isolated features [1,2]. The input image is standardized through preprocessing and passed through a shared ResNet-50 encoder for hierarchical feature extraction. Multi-scale feature integration is performed using a Feature Pyramid Network neck Concatenation-Based Multi-Scale feature fusion (C-MSFF) to combine lower-level spatial information with higher-level semantic morphology cues. From the shared representation, a segmentation decoder predicts fragmented regions at the pixel level, while two classification heads predict developmental stage and blastomere symmetry as embryo-level outputs. This shared representation with task-specific heads follows the general principle of multi-task learning. Training uses a weighted multi-task objective to optimize segmentation and classification jointly, while accounting for task imbalance and sparse fragmentation foregrounds. Final evaluation is performed separately for segmentation and classification outputs using Dice/IoU-based overlap measures and class-wise metrics suited to imbalanced morphology labels.

A substantial manual annotation effort was conducted using LabelMe on more than 10,000 cleavage-stage embryo images, after which 9,137 eligible annotated t2/t4 embryo microscopy images from two IVF centers-primarily Taichung Veterans General Hospital, with additional asymmetric and discarded cases from Chung Shan Medical University Hospital for minority-class enrichment; were curated with pixel-level cytoplasmic fragmentation annotations and image-level developmental-stage and blastomere-symmetry labels. The dataset was divided into 7,309 training, 914 validation, and 914 held-out test images using an approximately 80/10/10 stratified split based on the joint stage–symmetry label to preserve the distribution of all six morphology combinations and reduce subgroup imbalance. The split was generated using the train_test_split function in scikit-learn with a deterministic random seed of 42 for reproducibility [27].

### 3.2. Data labeling

Each embryo image was labeled with three supervision targets: developmental cell stage, blastomere symmetry, and cytoplasmic fragmentation. The developmental-stage target encoded the visible cleavage state as t2 or t4, corresponding to two-cell and four-cell cleavage-stage embryos. Early cleavage timing has been associated with later blastocyst formation and embryo quality [6], while morphokinetic parameters have been linked with implantation-related outcomes [28,29]. In particular, t2 and t4 are established morphokinetic landmarks in embryo time-lapse annotation and provide developmental context for interpreting cleavage-stage morphology.

The symmetry target was encoded as symmetric, asymmetric, or discarded, reflecting the visual regularity of blastomere size and spatial arrangement. Uneven blastomere size has been associated with lower implantation and pregnancy rates [4], and Day 2 blastomere symmetry has been linked with later blastocyst grade and ploidy status [5]. Fragmentation was annotated as a pixel-level target because cytoplasmic fragmentation is associated with reduced implantation and pregnancy potential [3]. Pixel-level localization was used to make fragmentation assessment spatially inspectable. Polygon regions corresponding to cytoplasmic fragments were manually delineated and converted into binary masks, with fragmentation treated as foreground and all remaining pixels treated as background. The resulting binary mask served as the expert reference mask for the fragmentation segmentation branch, following supervised biomedical image segmentation using pixel-level reference annotations [15]. Images with incomplete fragmentation annotations or missing morphology labels were excluded from supervised training and evaluation to ensure that all included samples contained complete supervision for the three EMBRACE outputs.

In this study, the symmetry evaluation method proposed by Yang et al. [14] was adopted and extended to define degree-of-symmetry categories for individual cleavage-stage embryo images. In that approach, the instantaneous symmetry score, denoted as $s_{\text{sym}}^{CS}(i)$, is calculated for a single frame during a given cleavage stage, such as cs2 or cs4. The score is defined as the ratio between the standard deviation of blastomere areas and the mean blastomere area within the same frame:

$$s_{\text{sym}}^{CS}(i) = \frac{\sigma_A^{CS}(i)}{\mu_A^{CS}(i)}, \tag{1}$$

where $\sigma_A^{CS}(i)$ and $\mu_A^{CS}(i)$ denote the standard deviation and mean area of blastomeres in frame i at cleavage stage CS, respectively. A higher $s_{\text{sym}}^{CS}(i)$ indicates greater variation in blastomere size and therefore a lower degree of symmetry. Yang et al. further defined a stage-level symmetry score, denoted as $s_{\text{sym}}^{CS}(i)$, by averaging the instantaneous symmetry scores across all frames belonging to the same cleavage stage:

$$S_{sym}^{CS} = \frac{1}{|I_{CS}|} \sum_{i \in I_{CS}} s_{sym}^{CS}(i), \quad (2)$$

where $I_{CS}$ represents the set of frame indices belonging to the corresponding cleavage stage. In their statistical analysis of 14,378 embryos, Yang et al. reported stage-level symmetry scores of $S_{\text{sym}}^{cs2} = 0.12 \pm 0.10$ and $S_{\text{sym}}^{cs4} = 0.20 \pm 0.09$.

Based on these reported distributions, the present study extended the stage-level symmetry-score thresholds to single-image symmetry categorization. Because the current dataset consists of static cleavage-stage images rather than complete time-lapse sequences, the instantaneous symmetry score was used to assign each image into one of three degree-of-symmetry categories:

$$\text{DoS Categories:} \quad \begin{cases} Symmetric, & s_{sym}{}^{cs2}(i) \in [0,0.12] \quad or \quad s_{sym}{}^{cs4}(i) \in [0,0.2] \\ Discarded, & s_{sym}{}^{cs2}(i) \in (0.12,0.22] \quad or \quad s_{sym}{}^{cs4}(i) \in (0.2,0.29] \\ Asymmetric, & s_{sym}{}^{cs2}(i) > 0.22 \quad or \quad s_{sym}{}^{cs4}(i) > 0.29 \end{cases} \quad (3)$$

The symmetric category represents embryos with relatively uniform blastomere areas within the reported mean-score range. The discarded category represents intermediate or borderline cases between the mean and approximately one standard deviation above the mean, where visual ambiguity may reduce label reliability. The asymmetric category represents embryos with symmetry scores greater than one standard deviation above the reported mean, indicating more pronounced blastomere-size variation. This thresholding strategy was used to support consistent label assignment for symmetry-related model training while preserving a separate discarded category for ambiguous or lower-quality cases. Note that for data recording and future research needs, samples were also indexed by cs2-symmetric, cs2-asymmetric, cs4-symmetric, and cs4-asymmetric. Herein, for the symmetry-grading task at hand, the final image-level labels were kept as three categories: symmetric, asymmetric, and discarded, consistent with the model output.

### 3.3. Data augmentation

The private and de-identified EmbryoScope™ dataset used in this study was collected from the Department of Obstetrics and Gynecology at Taichung Veterans General Hospital (VGHTC) between 2021 and 2023. The EmbryoScope™ dataset contains TLI image data and cleavage timing data.

The raw TLI images were obtained from TCVGH and CSMUH with differing spatial resolutions of 500 x 500 and 800 x 800 pixels, respectively. To ensure spatial consistency across data sources, all images were resampled to a standardized resolution of 512 x 512 pixels. Subsequently, the dataset was annotated using the custom developed by the Labelme. All embryo images were resized to 299 × 299 pixels using aspect-ratio-preserving padding to reduce geometric distortion while standardizing the network input size. Binary fragmentation masks were resized using nearest-neighbor interpolation to preserve discrete **foreground–background** labels and avoid interpolation-induced label artifacts. Images were normalized using ImageNet statistics to align with the initialization of the ImageNet-pretrained ResNet-50 encoder [23]. During training, spatial augmentations were applied jointly to images and masks so that geometric correspondence between embryo morphology and fragmentation annotations was preserved. Photometric augmentations were applied only to images because intensity and contrast variation can improve robustness to imaging differences, whereas applying such transformations to binary masks would corrupt the reference labels. These preprocessing and augmentation steps were designed to standardize image geometry, preserve mask integrity, and improve model robustness while maintaining correspondence between embryo images and fragmentation masks.

### 3.4 Architecture of EMBRACE

Fig.1 1 illustrates the architecture of EMBRACE. A preprocessed cleavage-stage embryo image of size 299×299 pixels is provided as input to a shared ImageNet-pretrained ResNet-50 encoder, which extracts hierarchical feature representations at multiple spatial resolutions. For cytoplasmic-fragmentation segmentation, intermediate and deep encoder features are integrated through the proposed Concatenation-Based Multi-Scale Feature Fusion (C-MSFF) module and subsequently processed by a refinement tower and U-Net-style decoder. In parallel, the deepest encoder representation is globally projected and supplied to two task-specific classification heads for t2/t4 developmental-stage prediction and three-class blastomere-symmetry grading. This shared-encoder, task-specific-branch design enables EMBRACE to jointly learn local, boundary-sensitive fragmentation features and global embryo-level morphological characteristics within a unified multi-task framework.

The segmentation Network branch receives the fused feature map Fs generated by the C-MSFF module and processes it through a convolutional refinement tower, two U-Net-style decoder blocks, and a segmentation prediction head. The refinement tower contains four consecutive 3×3 convolutional layers, each followed by 32-group normalization and ReLU activation, and preserves the 256-channel representation at approximately one-eighth of the input resolution (38×38). The first decoder block bilinearly upsamples the refined feature map to match the spatial resolution of the encoder feature L1(256×75×75), concatenates the two tensors along the channel dimension, and applies two 3×3 convolution–group-normalization–ReLU operations to produce a 128-channel feature map. The second decoder block similarly upsamples this representation to match the initial encoder feature S0(64×150×150), performs channel-wise concatenation, and applies two 3×3 convolutional operations to obtain a 64-channel decoder output. The segmentation head subsequently applies a 3×3 convolution that reduces the representation from 64 to 32 channels, followed by 16-group normalization and ReLU activation, and a final 1×1 convolution that maps the 32-channel feature vector at each spatial location to a single fragmentation logit. The 1×1 convolution therefore functions as a learned pixel-wise linear classifier without modifying the spatial dimensions. The resulting logit map is bilinearly resized to the original 299×299 input resolution. During inference, sigmoid activation converts the logits into foreground probabilities, which are thresholded at 0.5 to obtain the final binary fragmentation mask. This encoder–decoder design uses skip-connected high-resolution features to recover spatial information required for pixel-level fragmentation localization [15]. The described channel transitions and normalization layers correspond to the implemented architecture.

The classification component of EMBRACE uses the deepest shared encoder feature map, L4, which contains 2048 channels and encodes high-level semantic information about the entire embryo. Unlike fragmentation segmentation, which depends on fine-grained spatial and boundary information, developmental-stage classification and blastomere-symmetry grading require global interpretation of blastomere number, relative size, and spatial organization. Accordingly, L4 is used as the input to the global projection module and the two task-specific classification heads, enabling the model to capture embryo-level morphological characteristics relevant to t2/t4 stage prediction and symmetry grading.

The classification feature-extraction pathway begins with adaptive global average pooling, which transforms the Layer 4 feature map $L_4 \in R^{B\times 2048\times 10\times 10}$ into a 2048-dimensional vector for each image by averaging the activation of every channel across its 10×10 spatial grid. A fully connected layer introduces non-linearity. Dropout with a probability of p=0.3 is then applied to regularize the shared classification representation. During training, each element of the 256-dimensional feature vector is independently set to zero with a probability of 0.3, thereby reducing reliance on individual features and limiting overfitting. During inference, dropout is disabled, and the complete 256-dimensional feature vector is passed unchanged to the task-specific classification heads. This sequence constitutes the global projection module: adaptive global average pooling, flattening, a 2048→256 linear projection, ReLU activation, and dropout.The shared 256-dimensional classification feature vector is fed to two independent task-specific linear classifiers. The developmental-stage head applies a fully connected layer with a 256→2 mapping to generate two logits corresponding to the t2 and t4 classes, representing two-cell and four-cell cleavage-stage embryos, respectively.

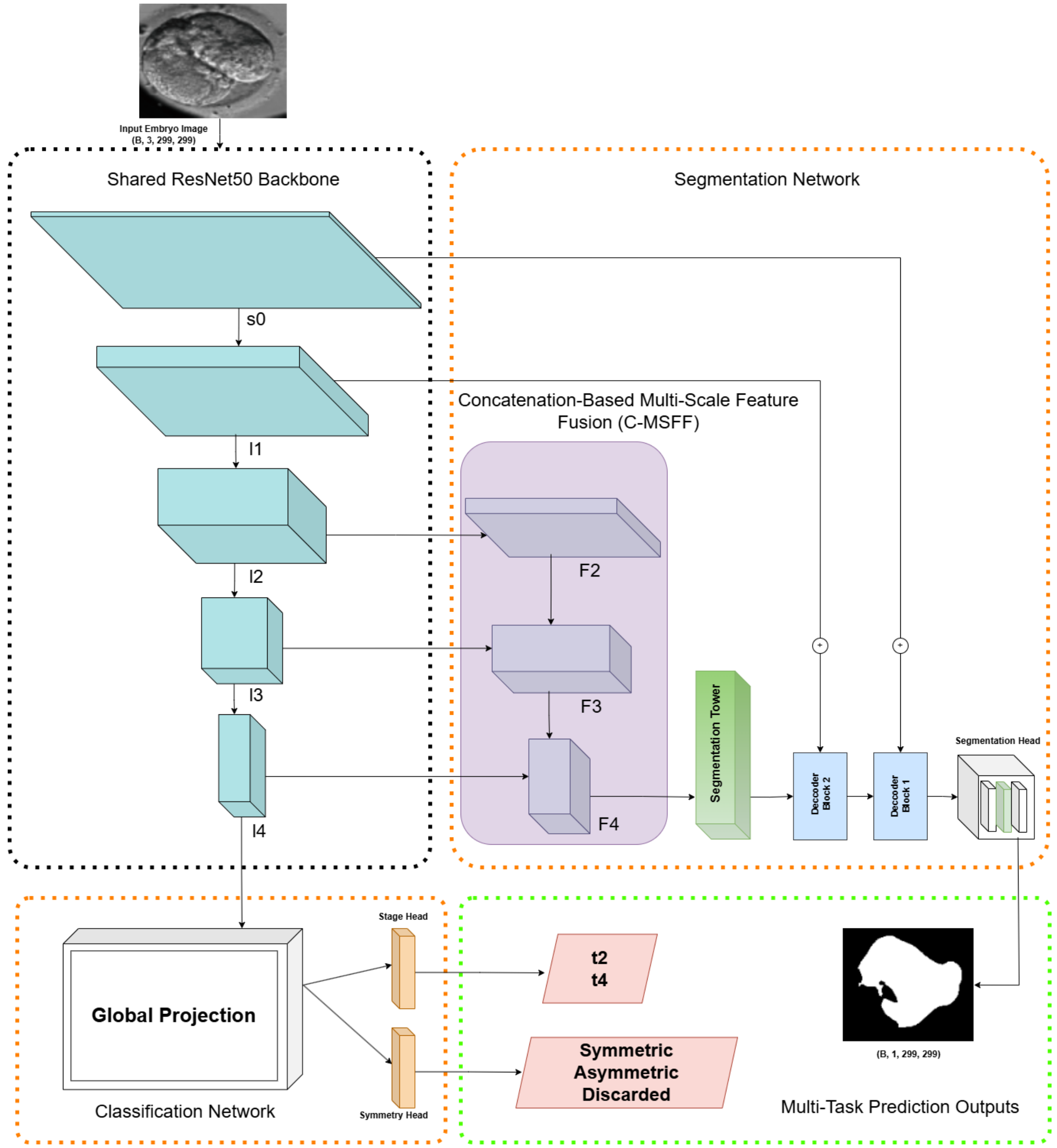

Figure 1. EMBRACE framework. The model receives a preprocessed cleavage-stage embryo image, extracts shared features through a ResNet-50/ Concatenation-Based Multi-Scale Feature Fusion (C-MSFF), and generates three task-specific outputs: a fragmentation mask, a developmental-stage prediction, and a blastomere symmetry grade.

The blastomere-symmetry head independently applies a 256→3 linear transformation to generate logits for the symmetric, asymmetric, and discarded categories. The two heads contain separately learned weights and biases, allowing each task to establish its own decision boundaries while retaining the common embryo-level representation learned by the shared encoder and global projection module. During inference, the class associated with the maximum logit is selected independently for each task. The implemented architecture therefore produces developmental-stage logits of shape B×2 and symmetry logits of shape B×3.

### 3.4.1 Fragmentation Segmentation Branch

The fragmentation segmentation branch converts the fused C-MSFF representation into a pixel-wise binary fragmentation mask through a convolutional refinement tower, two U-Net-style decoder blocks, and a segmentation prediction head. The fused feature map, $F_s \in R^{B\times256\times38\times38}$, is first processed by four consecutive blocks, each comprising a 3×3 convolution, group normalization with 32 groups, and ReLU activation. This refinement tower preserves both the 256-channel dimensionality and the 38×38 spatial resolution while adapting the multi-scale representation to dense fragmentation prediction. The first decoder block bilinearly upsamples the refined feature map to 75×75, concatenates it with the 256-channel L1 encoder feature, and applies two 3×3 convolution–group-normalization–ReLU operations to produce a 128-channel feature map. The second decoder block upsamples this representation to 150×150, concatenates it with the 64-channel S0 encoder feature, and applies two additional convolutional refinement operations to generate a 64-channel output. These skip connections restore higher-resolution spatial information that is reduced during encoder downsampling. The segmentation prediction head then applies a 3×3 convolution that reduces the representation from 64 to 32 channels, followed by group normalization with 16 groups and ReLU activation. A final 1×1 convolution maps the 32-channel feature vector at each spatial location to a single fragmentation logit, thereby functioning as a learned pixel-wise linear classifier without altering the spatial dimensions. The logit map is subsequently resized by bilinear interpolation to the original 299×299 input resolution. During inference, sigmoid activation converts the logits into foreground probabilities, and a threshold of 0.5 produces the final binary fragmentation mask. The network output has shape B×1×299×299, or 1×299×299 for an individual image.

### 3.4.2 Developmental-Stage and Degree of Symmetry Classification Heads

The developmental-stage and blastomere-symmetry classification tasks share a common global feature-extraction pathway before diverging into separate task-specific output layers. The deepest encoder feature map, $L_4 \in R^{B\times2048\times10\times10}$, is first processed by adaptive global average pooling, which averages each channel over its spatial dimensions and produces a 2048-dimensional feature vector for each input image. This vector is flattened and projected through a fully connected layer from 2048 to 256 dimensions, followed by ReLU activation and dropout with p=0.3, yielding a shared classification representation $z \in R^{B\times256}$. The developmental-stage head applies an independent linear transformation from 256 to 2 dimensions to generate logits for the t2 and t4 classes. In parallel, the blastomere-symmetry head applies a separate linear transformation from 256 to 3 dimensions to generate logits for the symmetric, asymmetric, and discarded categories. Because the two output layers have independently learned weights and biases, they can establish task-specific decision boundaries while sharing the upstream encoder and global projection module. During inference, the class with the largest logit is selected independently for each task. The resulting output shapes are B×2 for developmental-stage classification and B×3 for blastomere-symmetry grading.

## 3.5. Loss function, training protocol, and inference pipeline

The model was optimized using a weighted loss function to jointly train the fragmentation segmentation branch, developmental-stage classification head, and blastomere symmetry grading head. Multi-task learning allows related prediction tasks to share representations. Still, the loss formulation must be carefully balanced because dense segmentation and image-level classification may impose different optimization demands on the shared backbone. The total loss function was defined as:

$$L_total = \alpha * L_seg + \beta * L_stage + \gamma * L_sym, \quad (4)$$

where *L_seg, L_stage, and L_sym* represent the fragmentation segmentation loss, developmental-stage classification loss, and blastomere symmetry classification loss, respectively. The weighting coefficients $\alpha$, $\beta$, and $\gamma$ control the relative contribution of each task during joint optimization. In this study, these coefficients were empirically set as $\alpha$=1.0, $\beta$=0.5 and $\gamma$=0.5, respectively, to prioritize fragmentation segmentation while maintaining balanced supervision from the two classification tasks.

The segmentation loss combined weighted binary cross-entropy, Dice loss, focal loss, Lovász loss, and boundary loss. Weighted binary cross-entropy was used to address severe foreground–background imbalance in fragmentation masks, because cytoplasmic fragments occupy a small proportion of the image foreground. Dice loss was used to optimize overlap-sensitive segmentation behavior, which is important for sparse biomedical segmentation targets [16]. Focal loss was used to reduce the dominance of easy pixels and emphasize difficult fragmentation pixels. Lovász loss was included as an intersection-over-union-aligned surrogate for mask optimization. Boundary loss was used to encourage better contour precision around visually ambiguous fragment boundaries in highly imbalanced segmentation settings [18]. The developmental-stage head was optimized using focal cross-entropy to reduce the influence of visually easy examples, whereas the symmetry head used inverse-frequency weighted cross-entropy to address class imbalance across symmetric, asymmetric, and discarded categories. Each component of the compound segmentation loss is summarized in Table 1.

Training was conducted for 100 epochs with mixed-precision to improve computational efficiency on the 32 GB GPU. Model selection used a composite validation score prioritizing segmentation Dice, followed by symmetry balanced accuracy and developmental-stage F1. During inference, each preprocessed embryo image was processed once through the trained network to generate three outputs: a fragmentation probability map, a t2/t4 developmental-stage prediction, and a blastomere symmetry grade. The fragmentation probability map was thresholded at 0.5 to generate the binary fragmentation mask, while the classification heads used the highest softmax probability as the predicted class.

Table 1. Compound segmentation loss components and their rationale.

| Failure mode | Loss component | Purpose |
|---|---|---|
| Foreground fragments dominated by background | Weighted BCE | Up-weights positive fragmentation pixels under sparse foreground conditions |
| Poor mask overlaps despite pixel-wise learning | Dice loss | Directly improves overlap-sensitive segmentation behavior |
| Easy pixels dominate gradient | Focal loss | Down-weights easy pixels and emphasizes difficult fragmentation pixels |
| Dice provides limited signal at very low overlap | Lovasz loss | Provides IoU-aligned surrogate for mask optimization |
| Boundary ambiguity around fragments | Boundary loss | Emphasizes contour and edge near fragmented regions |

### 3.6. Evaluation metrics and statistical analysis

Fragmentation segmentation was evaluated using Dice coefficient and intersection over union (IoU/Jaccard) as the primary overlap metrics. Cytoplasmic fragmentation often occupies a small foreground region, making pixel accuracy potentially misleading under severe foreground-background imbalance. Dice coefficient measures overlap between predicted and reference masks, while IoU/Jaccard provides a stricter region-overlap measure aligned with segmentation quality. Precision and recall were additionally computed to distinguish over-segmentation from missed fragmented regions. Qualitative overlays were reviewed by comparing the input embryo image, expert reference mask, predicted mask, and overlay visualization, with particular attention to low-contrast or ambiguous fragment boundaries.

Developmental-stage classification was evaluated using accuracy, macro-F1, receiver operating characteristic area under the curve (AUC), and the confusion matrix, with standard metric computation implemented using scikit-learn. These metrics were used to assess both overall t2/t4 discrimination and class-wise performance. Blastomere symmetry grading was evaluated using balanced accuracy, macro-F1, class-specific F1-scores, quadratic weighted kappa, and the confusion matrix. Balanced and class-wise metrics were emphasized because symmetry labels were imbalanced, particularly for minority morphology categories. EXP1-EXP6 were compared under the same dataset split, preprocessing pipeline, and held-out test protocol to ensure that performance differences reflected model configuration rather than changes in data handling. Confidence intervals and repeated-run analyses were not performed in the present study and should be included in future validation work.

## 4. Experimental Results

### 4.1. Dataset characteristics and split distribution

The final supervised dataset included 9,137 annotated cleavage-stage embryo images. The dataset was divided into 7,309 training images, 914 validation images, and 914 held-out test images using an 80/10/10 stratified split. The split preserved the joint distribution of developmental stage and blastomere symmetry across all six stage–symmetry groups: t2-asymmetric, t2-discarded, t2-symmetric, t4-asymmetric, t4-discarded, and t4-symmetric. This stratification was important because class imbalance can influence model training and performance estimation, particularly when minority categories are underrepresented. Symmetric embryos formed the largest category in each partition, whereas asymmetric embryos were the smallest group, especially the t4-asymmetric subgroup. This distribution motivated the use of class-aware sampling and loss weighting during model training. The source-specific image and annotation distribution is reported in Table 2, and the final stratified train/validation/test split used for model development and held-out evaluation is shown in Table 3

Table 2. Distribution of TCVGH and CSMUH cleavage-stage embryo images and annotations

| **Hospital** | **Data** | **T2 Sym** | **T2 Asym** | **T4 Sym** | **T4 Asym** | **T2 Discarded** | **T4 Discarded** | **Total** |
|---|---|---|---|---|---|---|---|---|
| **TCVGH** | Images | 3862 | **80** | 3700 | **216** | 851 | 569 | 9,278 |
| | **Json** | **3434** | **68** | **1411** | **182** | **182** | **506** | **6,642** |
| **CSMUH** | Images | 0 | **627** | 0 | **214** | 2,936 | 1323 | 5100 |
| | **Json** | **0** | **423** | **0** | **186** | **186** | **986** | **2,499** |

Table 3. Dataset split and joint stage–symmetry distribution.

| Split | Total | T2-Asym | T2-Discarded | T2-Sym | T4-Asym | T4-Discarded | T4-Sym |
|---|---|---|---|---|---|---|---|
| **Train** | 7,309 | 393 | 1,314 | 2,986 | 294 | 1,194 | 1,128 |
| **Validation** | 914 | 49 | 164 | 374 | 37 | 149 | 141 |
| **Test** | 914 | 49 | 164 | 374 | 37 | 149 | 141 |

### 4.2. Test performance

EXP-1 was selected as the final EMBRACE configuration because it provided the strongest overall balance across the three prediction tasks, particularly by preserving the highest fragmentation Dice while maintaining strong developmental-stage and blastomere symmetry performance. On the held-out test set of 914 images, EMBRACE achieved Dice = 0.781 and IoU = 0.677 for fragmentation segmentation. These overlap-based metrics are appropriate for segmentation tasks with sparse foreground regions, where pixel accuracy can be misleading [16]. For developmental-stage classification, the model achieved accuracy = 0.995, macro-F1 = 0.994, and AUC = 1.000. For blastomere symmetry grading, the model achieved balanced accuracy = 0.901, macro-F1 = 0.907, and quadratic weighted kappa = 0.859. Balanced and class-wise metrics were emphasized because the dataset contained imbalanced morphology categories. A concise summary of the final test performance across the three EMBRACE outputs is summarized in Table 4.

Table 4. Overall test performance of EMBRACE across segmentation and classification tasks.

| Task | Primary metric(s) | Result |
|---|---|---|
| Fragmentation segmentation | Dice; IoU | 0.781; 0.677 |
| Stage classification | Accuracy; macro-F1; AUC | 0.995; 0.994; 1.000 |
| Symmetry grading | Balanced accuracy; macro-F1; quadratic weighted kappa | 0.901; 0.907; 0.859 |

Fig. 2 shows examples of cytoplasmic fragmentation segmentation. Fig. 2(a) and Fig. 2(b) show successful high-overlap cases in which the predicted masks closely matched the expert annotations, particularly when fragmented regions were visually distinct and sufficiently separated from the surrounding blastomeres. These examples indicate that EMBRACE can localize clear cytoplasmic fragments with good spatial agreement. In contrast, Fig. 2(c)-(f) show challenging and failure cases involving boundary deviation, small or low-contrast fragments, locally scattered fragmentation, and ambiguous fragment borders. In these cases, the model partially underestimated or missed subtle fragmented regions, reflecting the difficulty of segmenting sparse foreground structures against visually similar cytoplasmic texture. These qualitative findings are consistent with the quantitative Dice and IoU results and explain why fragmentation segmentation remained the most challenging output among the three EMBRACE tasks.

Overall the qualitative results suggest that EMBRACE performs reliably when fragmentation regions are visually clear, sufficiently large, and spatially distinguishable from blastomeres. However, segmentation accuracy decreases when fragments are small, low contrast, irregularly distributed, or located near ambiguous cytoplasmic boundaries. This observation supports the need for boundary-sensitive and imbalance-aware optimization in cleavage-stage embryo fragmentation segmentation.

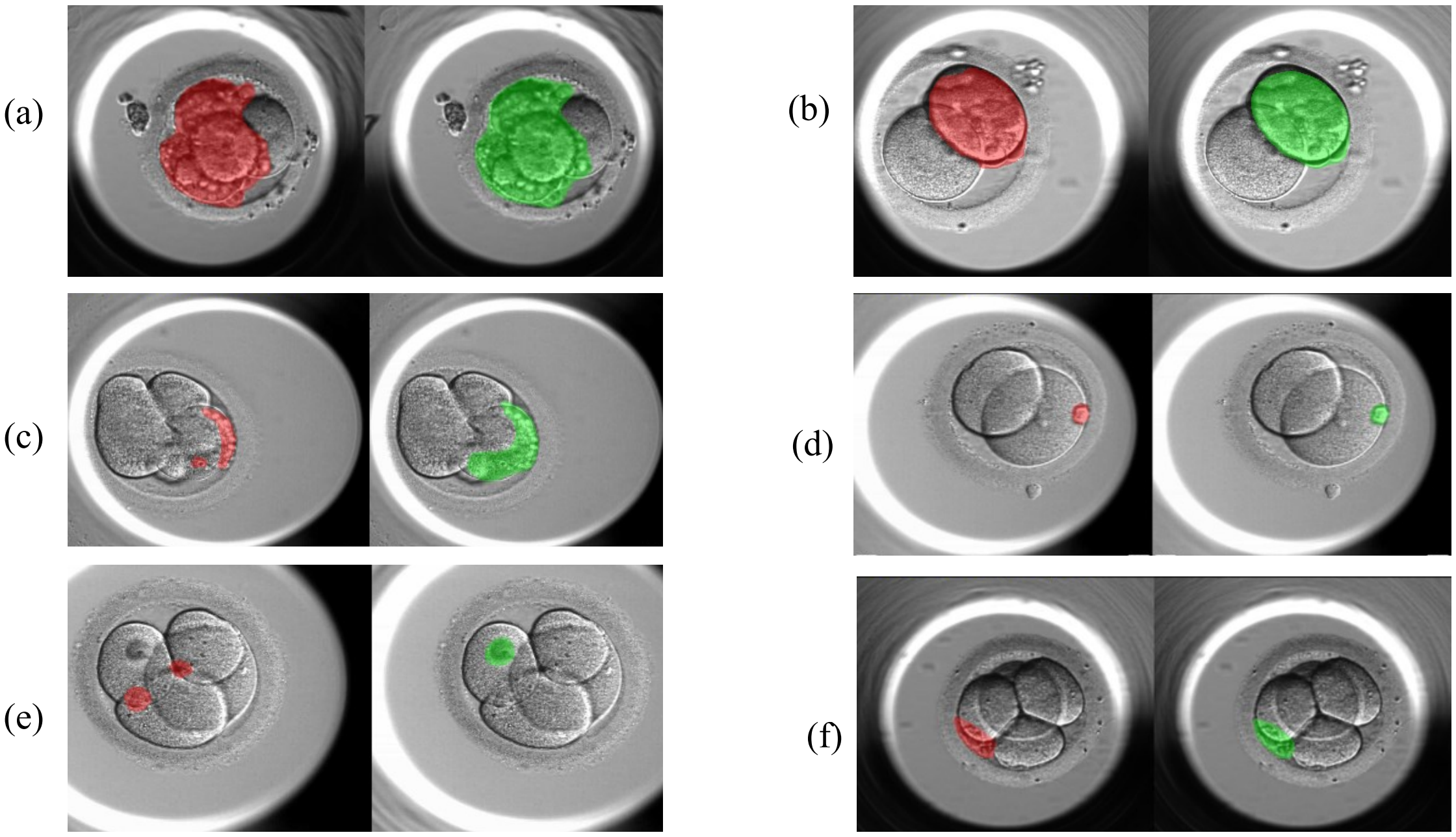

Figure 2. Cytoplasmic fragmentation segmentation. (a) large fragmented region. (b) Successful segmentation. (c) Morphologically challenging case with boundary deviation. (d) Small or low-contrast fragmented region. (e) Borderline fragmentation with multiple local fragmented regions. (f) Failure case involving ambiguous fragment boundaries. In each panel, the left image shows the expert annotation (red region) and the right image shows the model prediction (green region).

Fig. 3(a) shows that developmental-stage classification achieved near-complete separation between t2 and t4 embryos. EMBRACE correctly classified 586 of 587 t2 embryos and 325 of 327 t4 embryos, resulting in only three stage-classification errors across 914 test images. This corresponded to accuracy = 0.995, macro-F1 = 0.994, and AUC = 1.000. The confusion matrix shows that one t2 embryo was predicted as t4 and two t4 embryos were predicted as t2. This near-saturated performance should be interpreted in the context of the current binary t2/t4 classification setting, where visible two-cell and four-cell morphologies are comparatively distinct.

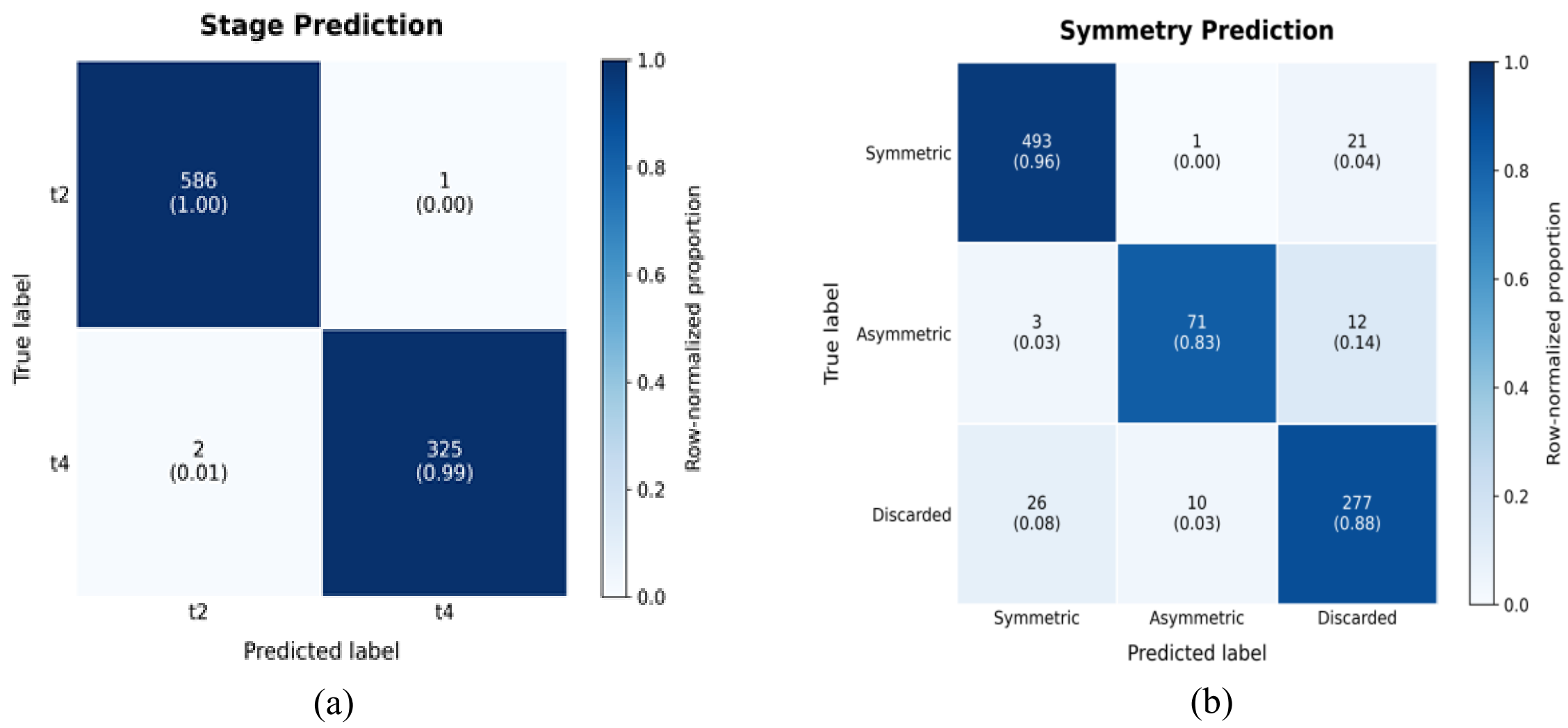

Figure 3. Confusion matrices (a) t2/t4 developmental-stage classification. (b) three-class blastomere symmetry grading.

Fig. 3(b) shows the confusion matrix for three-class blastomere symmetry grading. EMBRACE achieved balanced accuracy = 0.901, macro-F1 = 0.907, and quadratic weighted kappa = 0.859 on the held-out test set. The model correctly classified 493 of 515 symmetric embryos, 71 of 86 asymmetric embryos, and 277 of 313 discarded embryos. Most remaining errors occurred between the symmetric and discarded categories, including 26 discarded embryos predicted as symmetric and 21 symmetric embryos predicted as discarded. This error pattern suggests that borderline morphology between visually regular embryos and embryos assigned to the discarded category remained more challenging than the distinction between clearly symmetric and asymmetric embryos. Balanced accuracy and macro-F1 were therefore emphasized because symmetry categories were imbalanced [25], and class-wise evaluation is more informative than accuracy alone when minority morphology groups are underrepresented. Class-wise F1-scores were 0.949 for symmetric embryos, 0.881 for asymmetric embryos, and 0.892 for discarded embryos. These findings indicate that symmetry grading was reliable overall, but borderline symmetric–discarded cases remained the main source of residual error.

### 4.3. Ablation study

A controlled ablation study was conducted to evaluate the contribution of the main architectural and optimization components of the proposed EMBRACE framework. Six experimental configurations were compared using the same dataset split, preprocessing pipeline, training protocol, and held-out test evaluation to ensure fair comparison. All experiments used the 9,137-image dataset, comprising 7,309 training images, 914 validation images, and 914 held-out test images, with stratification by developmental stage and blastomere-symmetry label. Each configuration was developed from a common multi-task foundation, with specific modules introduced progressively to assess their effect on fragmentation segmentation, developmental-stage classification, and blastomere-symmetry grading. Model selection was based on a composite validation criterion prioritizing segmentation Dice, symmetry balanced accuracy, and developmental-stage F1, reflecting the multi-task objective of EMBRACE. Fragmentation segmentation was considered the primary spatial endpoint, whereas developmental-stage and symmetry classification were used to assess embryo-level morphology reliability.

#### EXP1: Baseline multi-task learning.

EXP1 served as the reference configuration for the ablation study. It used the compact EMBRACE baseline, consisting of a shared ResNet-50 encoder for hierarchical feature extraction [23], a multi-scale feature fusion representation, a U-Net-style decoder for pixel-wise fragmentation segmentation, and two independent classification heads for developmental-stage prediction and blastomere-symmetry grading. This shared-encoder and task-specific-head design allowed the model to generate three clinically relevant outputs from a single embryo image: a fragmentation mask, a t2/t4 developmental-stage prediction, and a blastomere-symmetry grade.

For optimization, EXP1 used a compound segmentation loss combining weighted binary cross-entropy, Dice, focal, Lovász-type, and boundary-sensitive terms to address foreground–background imbalance, overlap accuracy, difficult pixels, IoU-oriented optimization, and ambiguous fragment contours. Developmental-stage classification was optimized using focal cross-entropy, whereas symmetry grading used weighted cross-entropy to reduce the effect of class imbalance. The overall objective used fixed task weights, with the largest contribution assigned to fragmentation segmentation because it was the primary spatial endpoint.

This configuration did not include Siamese consistency, uncertainty-based task weighting, cross-task attention, ordinal regression, domain-adversarial learning, ASPP, or deep supervision. Therefore, EXP1 provided a compact and controlled baseline for assessing whether the additional modules introduced in subsequent experiments produced meaningful improvements over standard multi-task learning.

**EXP2: Siamese consistency and uncertainty-based task weighting.**

EXP2 extended the EXP1 baseline by introducing Siamese consistency regularization and uncertainty-based task weighting. During training, two independently augmented views of the same embryo image were passed through the shared network, and a consistency term was used to encourage stable segmentation and classification outputs across paired views [30]. This design was motivated by the assumption that clinically relevant embryo morphology labels should remain stable under reasonable image-level augmentation. At inference, the model retained the same single-image prediction structure as EXP1, ensuring that the additional Siamese branch affected only training and did not increase deployment complexity.

In addition, the fixed task weights used in EXP1 were replaced with learnable uncertainty-based weights to adaptively balance fragmentation segmentation, developmental-stage classification, and blastomere-symmetry grading losses. This was intended to reduce manual task-weight tuning and account for different convergence rates across dense segmentation and embryo-level classification objectives. Therefore, EXP2 evaluated whether consistency regularization and adaptive task balancing could improve the stability and reliability of the baseline multi-task model without changing its output structure.

**EXP3: Cross-task attention and CORAL ordinal regression.**

EXP3 extended the Siamese and uncertainty-weighted configuration by introducing cross-task attention and CORAL-based ordinal symmetry grading. Cross-task attention used the predicted fragmentation map as spatial guidance for embryo-level classification, allowing the stage and symmetry heads to incorporate fragmentation-related spatial evidence without modifying the segmentation output [31]. This design was motivated by the clinical relationship between cytoplasmic fragmentation, blastomere organization, and overall embryo morphology.

In addition, the blastomere-symmetry head was reformulated using CORAL ordinal regression [33]. Instead of treating symmetric, asymmetric, and discarded embryos as unrelated nominal classes, CORAL modelled these categories as ordered morphology states with increasing abnormality. This was intended to better reflect the clinical structure of embryo-quality assessment.

EXP3 therefore evaluated whether explicit task interaction and ordinal morphology modelling could improve embryo-level classification while preserving fragmentation segmentation as the primary spatial endpoint. The model retained Siamese consistency regularization and uncertainty-based task weighting from EXP2, allowing the additional effect of cross-task attention and ordinal grading to be assessed within the stabilized multi-task setting.

**EXP4: Domain-adversarial learning.**

EXP4 evaluated the contribution of domain-adversarial learning within the Siamese and uncertainty-weighted configuration. In contrast to EXP3, cross-task attention and CORAL ordinal regression were excluded, allowing the effect of domain adaptation to be examined separately. A domain discriminator was connected to the shared encoder through a gradient reversal layer, encouraging the encoder to reduce hospital- or acquisition-specific information in the learned representation.

This experiment was motivated by the multi-center nature of the embryo image dataset, where images may vary across hospitals because of differences in illumination, contrast, microscope settings, or acquisition conditions. By penalizing hospital-discriminative features during training, EXP4 tested whether domain-adversarial regularization could improve robustness across clinical imaging sources.

The optimization objective retained Siamese consistency regularization and uncertainty-based task weighting from EXP2. An additional domain-adversarial loss was introduced, while symmetry grading continued to use weighted cross-entropy rather than CORAL ordinal regression. Therefore, EXP4 specifically assessed whether cross-site regularization improved multi-task performance without introducing explicit task-attention or ordinal-grading mechanisms.

**EXP5: Combined auxiliary multi-task configuration.**

EXP5 combined the main auxiliary strategies evaluated in EXP2–EXP4 into a single multi-task configuration. This model included Siamese consistency regularization [30], uncertainty-based task weighting, cross-task attention, CORAL-based ordinal symmetry grading, and domain-adversarial learning. The purpose of this experiment was to determine whether these individually motivated components provided complementary benefits when integrated into one framework, or whether the additional objectives introduced optimization interference within the shared representation.

This configuration represented the highest-complexity auxiliary model before adding segmentation-specific architectural enhancements. It tested whether consistency regularization, adaptive task balancing, fragmentation-guided classification, ordinal morphology modelling, and cross-site regularization could jointly improve fragmentation segmentation, developmental-stage prediction, and blastomere-symmetry grading. Because multi-task learning can be affected by task compatibility and gradient conflict [20], EXP5 was also used to examine whether increasing auxiliary supervision improved overall performance or created trade-offs between dense segmentation and embryo-level classification.

The optimization objective retained uncertainty-weighted segmentation and developmental-stage losses, CORAL-based symmetry loss, Siamese consistency regularization, and domain-adversarial loss. Training was extended to support convergence under the more complex objective, and per-hospital evaluation was used to examine cross-site behavior.

**EXP6: Enhanced architecture with ASPP and deep supervision.**

EXP6 extended the combined auxiliary configuration by adding atrous spatial pyramid pooling (ASPP), deep supervision, and learning-rate warmup. ASPP was introduced to strengthen multi-scale contextual modelling for fragmented regions that may appear small, irregular, low contrast, or spatially dispersed [35]. An auxiliary segmentation head was added to provide intermediate decoder supervision, while learning-rate warmup was used to stabilize early optimization under the larger multi-loss objective.

This experiment was designed to determine whether segmentation-specific architectural enhancement could further improve the full multi-task framework. In particular, EXP6 tested whether additional contextual modelling and intermediate feature supervision improved fragmentation localization while preserving reliable developmental-stage classification and blastomere-symmetry grading. This was important because fragmentation segmentation was the primary spatial endpoint and represented the task most affected by boundary ambiguity, foreground sparsity, and local texture variation. The optimization objective retained uncertainty-based task weighting, Siamese consistency regularization, CORAL-based symmetry loss, and domain-adversarial regularization from the combined configuration. The auxiliary segmentation loss was gradually reduced during training so that early optimization benefited from intermediate supervision while final model selection remained focused on the main segmentation output. The quantitative comparison of EXP1–EXP6 is summarized in Table 5. EXP5 was reported using per-hospital evaluation, 0.711 for TCVGH and 0.761 for CSMUH in Dice outcome, respectively. Abbreviations: Acc, accuracy; Bal Acc, balanced accuracy; F1, F1-score; IoU, intersection over union; QWK, quadratic weighted kappa; Sym, symmetry.

Table 5. Ablation-study performance summary.

| Experiment | Dice | IoU | Stage Acc | Stage F1 | Sym Bal Acc | Sym Macro-F1 | QWK |
|---|---|---|---|---|---|---|---|
| **EMBRACE** | **0.781** | **0.677** | 0.995 | 0.994 | 0.901 | **0.907** | 0.859 |
| **EXP2** | 0.758 | 0.646 | 0.995 | 0.994 | **0.902** | 0.906 | 0.869 |
| **EXP3** | 0.772 | 0.664 | **0.997** | **0.996** | 0.876 | 0.872 | 0.873 |
| **EXP4** | 0.769 | NA | 0.996 | 0.995 | 0.891 | 0.901 | 0.841 |
| **EXP5** | 0.711 | NA | 0.999 | NA | 0.828 | NA | 0.777 |
| | 0.761 | NA | 0.988 | NA | 0.899 | NA | 0.836 |
| **EXP6(Enhanced)** | 0.764 | 0.653 | 0.992 | 0.9912 | 0.861 | 0.836 | 0.868 |

### Interpretation

**Segmentation performance.** EXP1 achieved the strongest fragmentation segmentation performance among all variants, indicating that the compact baseline was most effective for pixel-level fragment localization. In contrast, the additional modules introduced in EXP2-EXP6 consistently reduced segmentation performance. This pattern suggests that the main limitation was not insufficient model capacity, but the difficulty of learning sparse and low-contrast cytoplasmic fragmentation regions. Boundary ambiguity further makes this task difficult under highly imbalanced segmentation conditions. Additional objectives such as consistency regularization, domain alignment, ordinal learning, and auxiliary supervision may have increased optimization competition within the shared encoder. This interpretation is consistent with multi-task learning studies showing that task compatibility and gradient conflict can influence performance when multiple objectives share the same representation.

**Blastomere symmetry grading.** EXP1 provided the most reliable overall symmetry grading, while EXP2 showed only a marginal improvement in balanced accuracy, suggesting a limited benefit from view-invariant consistency learning. In contrast, models using CORAL ordinal regression showed weaker symmetry performance, particularly for the asymmetric class. This suggests that although symmetry categories have an ordered clinical meaning, the visual boundary between symmetric, asymmetric, and discarded embryos is not strictly linear. The asymmetric class represents a heterogeneous middle category, and ordinal constraints may restrict its decision space too strongly. Weighted cross-entropy in EXP1 provided more flexible class boundaries, which was more suitable for this imbalanced grading task.

**Ordinal consistency.** CORAL was used to model symmetry grading as an ordinal prediction task. In EXP3 and EXP5, ordinal consistency improved despite a reduction in overall symmetry classification performance. This suggests that ordinal modelling helped reduce severe symmetry-order errors by making predictions more likely to fall between neighboring categories rather than clinically distant classes. However, this improvement in ordinal consistency did not translate into better balanced accuracy or macro-F1, particularly for the asymmetric class. Therefore, CORAL was useful for preserving the ordered structure of symmetry grading, but it was not selected for the final model because the main objective required reliable class-wise discrimination across all symmetry categories.

**Domain adaptation.** EXP4 used domain-adversarial learning to encourage hospital-invariant feature representations through adversarial domain alignment [34]. In the ablation results, DANN maintained competitive embryo-level classification but did not preserve the primary fragmentation segmentation endpoint. This was useful for testing robustness across clinical sources, especially because imaging conditions and class distributions differed between centers. A likely explanation is that removing site-specific appearance information may also suppress subtle contrast, texture, or boundary cues that are useful for detecting small fragmented regions. Therefore, domain adaptation provided valuable robustness insight, but it was not selected for the final configuration because it weakened the primary spatial endpoint.

**Enhanced architecture.** EXP6 added atrous spatial pyramid pooling to capture multi-scale contextual information and deep supervision to strengthen intermediate feature learning [36]. In the ablation results, adding these components did not improve the overall multi-task balance compared with EXP1. This suggests that the main bottleneck was not insufficient receptive field or model depth, but the intrinsic difficulty of the data, including sparse cytoplasmic fragmentation, ambiguous segmentation boundaries, and class imbalance. Additional objectives may also increase optimization burden when task compatibility is limited, and gradient conflict can further affect shared multi-task representations. Therefore, EXP6 confirmed that greater complexity was not beneficial for this task, supporting the selection of the compact EXP1 configuration.

**Clinical and Technical Relevance:** The ablation findings suggest that, for cleavage-stage embryo morphology assessment, carefully designed supervision was more important than increasing architectural complexity. The compact EXP1 configuration benefited from imbalance-aware segmentation and classification losses and boundary-sensitive supervision for ambiguous fragmentation regions. This helped preserve pixel-level fragmentation sensitivity while maintaining reliable embryo-level predictions. From a clinical AI perspective, this supports the use of simpler and more stable models that are easier to validate, interpret, and maintain, particularly when the model output includes spatially inspectable segmentation evidence.

The additional modules still provided useful technical insights. Cross-task attention [31] was used to model task interaction between segmentation and classification features. Ordinal modelling was used to preserve the ordered structure of symmetry grading. Domain-adversarial learning offered a way to examine cross-site robustness through domain-invariant representation learning [34]. However, these targeted benefits did not consistently improve all endpoints, especially fragmentation segmentation. This pattern is consistent with the broader multi-task learning concern that task compatibility must be evaluated carefully, and gradient conflict can occur when dense segmentation and image-level classification objectives compete within a shared representation. Therefore, such modules may be useful in specific deployment settings, but the final EMBRACE configuration prioritized the compact baseline because it provided the strongest overall balance.

In summary, the ablation study showed that EXP1 provided the strongest overall balance across fragmentation segmentation, developmental-stage classification, and blastomere symmetry grading.

Although later experiments introduced targeted modules, including cross-task attention for task interaction, CORAL ordinal modelling for ordered symmetry prediction [33], and domain-adversarial learning for cross-site robustness, these gains did not translate into consistent multi-task improvement. In particular, added complexity often increased optimization competition and weakened the primary fragmentation segmentation endpoint. These findings are consistent with the claimed task compatibility and gradient conflict, and indicate that cleavage-stage embryo morphology assessment benefits more from a compact, stable, and well-regularized multi-task design than from combining multiple auxiliary modules. Cytoplasmic fragmentation regions are visually challenging, boundary ambiguity makes segmentation difficult, and class imbalance affects model learning and evaluation. Therefore, reliable supervision and class-imbalance handling were more important than increasing model complexity, supporting the selection of EXP1 as the final EMBRACE configuration. The essential training hyperparameters used for the final configuration are given in Table 6 to support reproducibility.

Table 6. Essential training hyperparameters

| **Hyperparameters** | **Specifications** |
|---|---|
| Input size | 299× 299 |
| Backbone | ImageNet-Pretrained ResNet-50 |
| Learning rate/ weight decay | $1\times10^{-4}$ / $1\times10^{-4}$ |
| Scheduler | Cosine annealing |
| Batch size/epochs | 8 /100 |
| Sampling | Weighted sampling by stage x symmetry labels |
| Split/seed | Stratified 80/10/10, seed 42 |
| Loss weights | $\alpha$=1.0, $\beta$=0.5 and $\gamma$=0.5 |
| Environment | PyTorch 2.1.2, CUDA 11.8, single 32 GB GPU, with AdamW optimizer |

### 4.4. Failure analysis

Qualitative review indicated that EMBRACE performed best when fragmentation boundaries were visually clear and fragmented regions were sufficiently separated from surrounding blastomere texture. Failure cases were mainly associated with borderline fragmentation, poor contrast, and ambiguous blastomere-fragment interfaces. These errors are consistent with the biological and visual difficulty of distinguishing small anucleate cytoplasmic fragments from surrounding embryo texture. Pixel-level mask review is important for visually inspecting segmentation behavior, while ambiguous boundaries and foreground-background imbalance remain known challenges in segmentation. The validation-score diagnostic analysis further indicated that segmentation contributed the majority of residual model error, while developmental-stage classification contributed the smallest share. This pattern is consistent with the result that segmentation is a dense, boundary-sensitive task, whereas t2/t4 classification involves visually distinct cleavage-stage categories. The task-level diagnostic contribution to the validation score is summarized in Table 7, reinforcing that segmentation was the dominant residual-error source.

Table 7. Validation-score diagnostic contribution by task.

| Task | Contribution to validation score | Interpretation |
|---|---|---|
| Segmentation | 0.110 (78%) | Main residual-error source; priority target for improvement |
| Symmetry | 0.030 (21%) | Secondary source of residual error |
| Stage | 0.001 (1%) | Near-saturated in current split |

### 4.5. Computational feasibility and inference time

EMBRACE generates cytoplasmic fragmentation segmentation, developmental-stage prediction, and blastomere symmetry grading in a single forward pass from one cleavage-stage embryo image. This single-pass design avoids maintaining separate segmentation and classification pipelines and may reduce the risk of inconsistent outputs across related morphology tasks. From a computational perspective, the compact EMBRACE configuration is also preferable to the auxiliary EXP2-EXP6 variants because it achieved the strongest overall task balance without adding Siamese branches, ordinal-regression modules, domain-adversarial components, ASPP, or deep supervision.

Qualitative ablation examples are shown in Fig. 4, where EMBRACE and EXP2-EXP6 are compared on the same representative embryo image. The visual comparison supports the quantitative findings in Table 5: the compact EMBRACE configuration produced more consistent fragmentation localization, whereas several auxiliary variants showed under-segmentation, over-segmentation, or boundary displacement. These observations suggest that added architectural complexity did not consistently improve the primary segmentation endpoint and may have increased optimization interference within the shared multi-task representation.

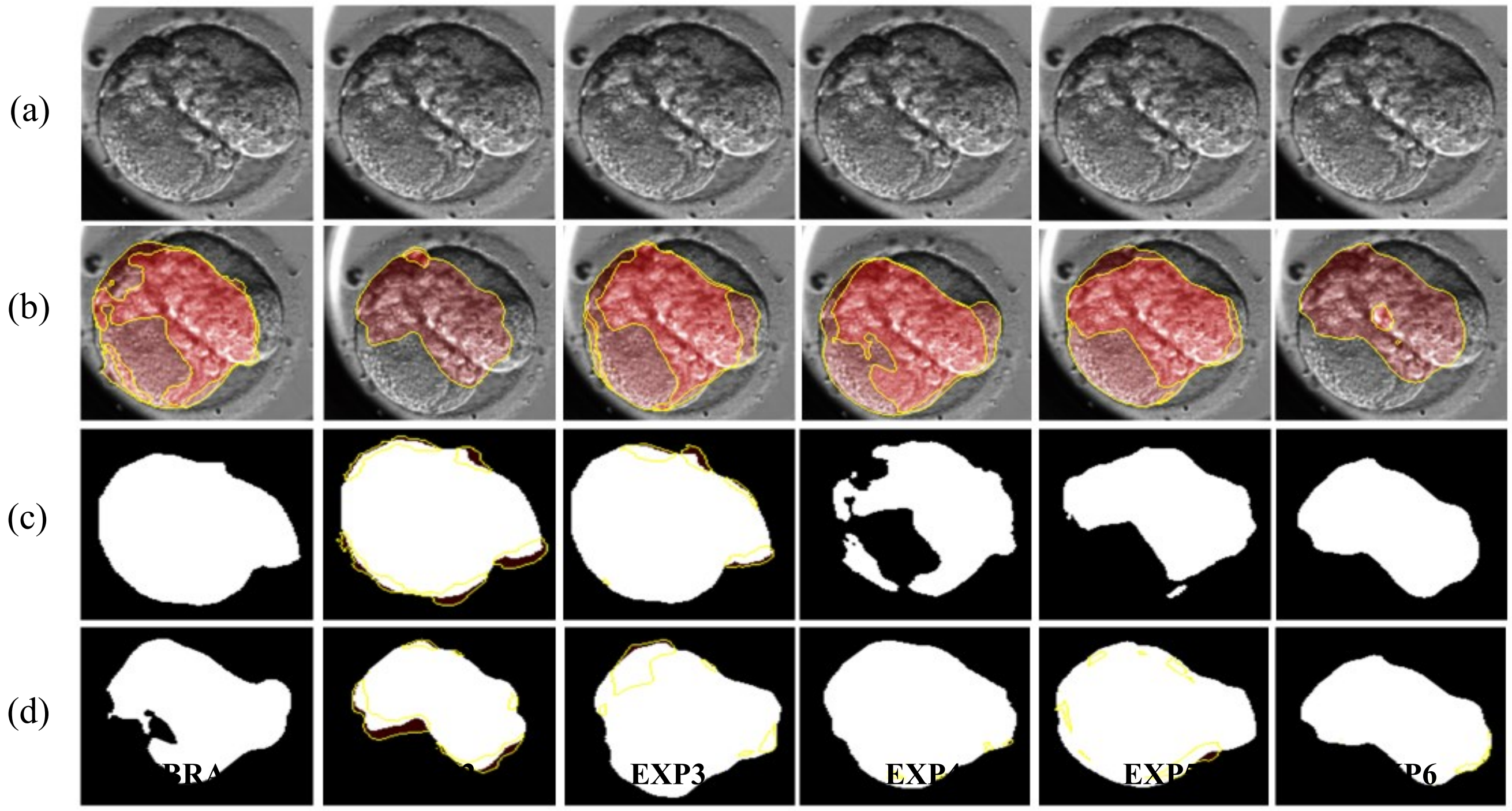

Figure 4. Qualitative visualization of ablation results for cytoplasmic fragmentation segmentation. Columns compare EMBRACE and EXP2-EXP6 on the same representative embryo image. Rows show (a) Input image, (b) Overlay visualization, (c) Expert mask, and (d) Predicted mask.

## 5. Discussion

This study introduced EMBRACE, a shared-encoder multi-task framework that jointly performs cytoplasmic fragmentation segmentation, t2/t4 developmental-stage classification, and blastomere symmetry grading from static cleavage-stage embryo microscopy images. This task design reflects the multi-feature structure of cleavage-stage embryo assessment, where fragmentation, developmental progression [6] and blastomere symmetry are interpreted together during morphology-based embryo evaluation [1]. The main finding is that the compact EXP-1 configuration achieved the strongest overall task balance, with high developmental-stage classification performance, clinically relevant blastomere symmetry grading, and spatially inspectable fragmentation masks. Pixel-level localization provides more interpretable evidence for fragmentation assessment than image-level scores alone. Segmentation remained the limiting task, reflecting the difficulty of localizing small, irregular, low-contrast fragmented regions. Ambiguous boundaries and foreground–background imbalance further increase the difficulty of boundary-sensitive segmentation.

The ablation results also suggest that more complex task-interaction strategies did not automatically improve performance in this dataset and task formulation. Variants incorporating Siamese consistency, uncertainty weighting, cross-task attention, domain-adversarial learning, ASPP, or deep supervision [36] did not outperform EXP1 on the main fragmentation segmentation endpoint. This observation is consistent with multi-task learning literature showing that related tasks can benefit from shared representation learning [19], but may also suffer when task compatibility is limited, gradient conflict occurs, or loss balancing becomes difficult. In this study, a restrained shared-encoder design with task-specific heads appeared to preserve segmentation performance more effectively than adding additional task-interaction modules, suggesting that architectural complexity should be justified empirically rather than assumed to improve multi-task embryo assessment.

The strong overall performance of EMBRACE can be attributed to three key architectural characteristics. First, the framework uses a shared ResNet-50 encoder with an Multi-Scale feature fusion representation that integrates fine-grained spatial information with high-level semantic morphology features, enabling simultaneous analysis of localized cytoplasmic fragmentation and global embryo structure. Second, EMBRACE adopts a clinically aligned multi-task learning strategy that jointly models fragmentation segmentation, developmental-stage classification, and blastomere symmetry grading, allowing the shared encoder to learn richer morphology-aware representations than single-task approaches. Third, the architecture balances shared representation learning with task-specific specialization through dedicated prediction heads, thereby reducing optimization interference between dense segmentation and image-level classification objectives. The ablation results further support this design choice, as several more complex variants failed to consistently outperform the compact baseline configuration. Collectively, these characteristics suggest that EMBRACE benefits not from architectural complexity alone, but from an effective balance between multi-scale feature integration, clinically relevant task coupling, and task-specific specialization.

Prior embryo AI studies have demonstrated the feasibility of computational embryo and oocyte assessment. Earlier work investigated embryo and oocyte classification using artificial intelligence techniques [11]. Static microscopy-based models have shown the feasibility of predicting embryo viability from optical microscopy images [12]. Time-lapse deep learning approaches have also been used to predict blastocyst formation and embryo quality from developmental image sequences [13]. These studies show that embryo microscopy images and time-lapse sequences contain visual information that can be modeled using machine learning and deep learning approaches.

However, many embryo AI models remain focused on image-level prediction, which limits their ability to provide spatial evidence for morphology-related decisions. This limitation is particularly important for cytoplasmic fragmentation because localized visual evidence is clinically meaningful in embryo assessment. Segmentation-based outputs can provide pixel-level spatial evidence that is more inspectable than image-level prediction alone. EMBRACE differs from these approaches by jointly modeling local fragmentation segmentation and global morphology classification from the same input image. This distinction is important because cleavage-stage assessment is not a single-label problem; embryologists interpret cytoplasmic fragmentation, developmental stage, and blastomere symmetry together during morphology-based embryo evaluation.

The proposed framework is also aligned with a broader direction in medical AI toward structured outputs that combine localization with clinically relevant prediction. In medical imaging, segmentation-based approaches are widely used to provide spatially interpretable outputs, while recent weakly supervised medical imaging literature has emphasized the importance of reducing annotation burden and improving clinically meaningful model interpretation [37]. Although the clinical domain differs, this broader literature supports the principle that medically useful AI systems should provide inspectable evidence rather than only an image-level prediction without spatial explanation. Direct numerical comparison across embryo AI studies should remain cautious because prior studies differ in task focus, including embryo/oocyte classification [11], static-image viability prediction, and time-lapse blastocyst prediction. They may also differ in datasets, imaging protocols, embryo stages, annotation procedures, label definitions, and evaluation endpoints.

The main clinical relevance of EMBRACE is its structured multi-output design. Instead of providing only a single embryo-level score, the framework produces a cytoplasmic fragmentation mask together with t2/t4 developmental-stage and blastomere symmetry predictions. This output structure is aligned with cleavage-stage embryo assessment, where embryologists consider cytoplasmic fragmentation, developmental progression, and blastomere regularity together during morphology-based evaluation.

The fragmentation mask may support embryology review by making the spatial basis of fragmentation assessment inspectable [14]. Because segmentation outputs can be compared directly with the embryo image and expert reference annotation, they provide pixel-level evidence for model review. This spatial output may also support auditability in research workflows, where predicted masks, expert annotations, and failure cases can be reviewed systematically to identify model behavior. Boundary-related errors should also be examined because ambiguous fragment boundaries remain challenging in imbalanced segmentation settings [18].

Technically, EMBRACE differs from many previous embryo AI approaches by combining spatially inspectable fragmentation segmentation with embryo-level morphology classification in a single framework. This structured output design is important because cleavage-stage assessment requires both localized evidence and global morphology interpretation. The ablation findings further indicate that a compact shared-encoder framework with task-specific heads was more stable than adding complex task-interaction modules, supporting the use of restrained and empirically validated multi-task designs for embryo image analysis.

Several limitations should be considered. First, this study used a retrospective dataset, and further external and prospective validation is required before generalizability or clinical readiness can be assumed. Although images from two IVF centers were included, the CSMUH cases were incorporated specifically to enrich underrepresented asymmetric and discarded cleavage-stage categories because these classes were limited in the primary TCVGH dataset. Therefore, the combined dataset should be interpreted as a class-imbalance enrichment strategy rather than as independent external validation. Center-wise performance and acquisition-specific effects were not fully evaluated. Robustness across different imaging devices, laboratory protocols, and center-specific image characteristics remains to be established in future embryo AI studies. Annotation practices may also affect model reliability because embryo morphology assessment [9] and time-lapse annotation can vary across observers.

Second, fragmentation annotation is inherently challenging because small cytoplasmic fragments may be visually similar to surrounding cytoplasmic texture or located near ambiguous blastomere boundaries. This can affect both the reference masks used for training and the estimated segmentation performance. Boundary uncertainty is particularly important for fragmentation segmentation, while sparse foreground regions can increase foreground-background imbalance. Therefore, future studies should report annotation procedures in greater detail, including annotator expertise, consensus rules, and inter-observer agreement. This is important because embryo morphology assessment can vary across observers [9], and annotation variability has also been reported in time-lapse embryo assessment [10].

Third, the dataset showed class imbalance, particularly for asymmetric and discarded embryos in the primary TCVGH cohort, as well as foreground sparsity in cytoplasmic fragmentation masks. To reduce developmental-stage and symmetry subgroup imbalance, additional asymmetric and discarded cleavage-stage cases from CSMUH were incorporated during dataset preparation. However, this enrichment strategy does not eliminate all imbalance-related concerns. Class imbalance can affect model optimization, metric interpretation, and the reliability of minority-class performance estimates. Sparse foreground annotations can also make fragmentation segmentation more sensitive to foreground–background imbalance. Therefore, performance for underrepresented categories, especially asymmetric embryos, should be interpreted cautiously and supported by class-wise metrics, confidence intervals, and center-wise subgroup analysis when available.

Fourth, the present framework used static cleavage-stage microscopy images and did not incorporate full time-lapse morphokinetic sequences. Time-lapse annotation provides standardized developmental landmarks such as t2 and t4. Early cleavage timing has been associated with blastocyst formation and embryo quality [6]. Morphokinetic parameters have also been linked with implantation-related outcomes. Therefore, the current model should be interpreted as a static-image morphology assessment framework rather than a full morphokinetic prediction system.

Fifth, the current evaluation does not establish implantation prediction, pregnancy prediction, live-birth prediction, or clinical deployment readiness. The model outputs are morphology-related predictions based on embryo image analysis, not validated reproductive outcome predictions. EMBRACE should therefore be interpreted as a research-oriented decision-support framework for embryo image analysis, consistent with the role of morphology-based embryo assessment in embryology practice. It should not be interpreted as a replacement for embryologists or as an autonomous embryo-selection system. Prior embryo AI studies have shown the feasibility of embryo/oocyte classification, static-image embryo viability prediction [12], and time-lapse blastocyst prediction, but clinical translation still requires external validation, prospective testing, workflow-level evaluation, calibration analysis, and clear reporting of failure modes before routine clinical use can be considered.

## 6. Conclusions

This study introduced EMBRACE, a unified multi-task deep learning framework for cleavage-stage embryo image analysis that jointly performs cytoplasmic fragmentation segmentation, t2/t4 developmental-stage classification, and blastomere symmetry grading. This task formulation reflects the multi-feature structure of cleavage-stage embryo assessment, where cytoplasmic fragmentation, developmental progression, and blastomere regularity are interpreted together during morphology-based evaluation. The framework combines spatially inspectable fragmentation masks with embryo-level morphology predictions, addressing a limitation of image-level embryo AI models such as embryo/oocyte classification, static-image viability prediction, and time-lapse blastocyst prediction, which may provide classification outputs without localizing the morphological evidence supporting the prediction. In the present evaluation, the EXP1 configuration achieved the strongest overall balance across segmentation, developmental-stage classification, and symmetry grading, supporting the feasibility of a structured multi-output approach for embryo morphology assessment.

The findings suggest that shared representation learning with task-specific heads can support coordinated local and global embryo morphology analysis, consistent with the general principle of multi-task learning. However, task compatibility, gradient conflict, and loss balancing should be considered when multiple objectives share the same encoder. The use of overlap-sensitive, imbalance-aware, and boundary-aware segmentation objectives was particularly relevant because cytoplasmic fragmentation can occupy a sparse foreground region. Dice-based overlap optimization supports sparse segmentation learning, Lovász-type objectives support IoU-oriented mask optimization, boundary-aware loss is relevant for ambiguous fragment boundaries, and imbalance-aware learning is important when foreground regions are limited. EMBRACE should therefore be interpreted as a research-oriented decision-support framework rather than an autonomous embryo-selection system. Clinical deployment requires further external and prospective validation, complete reporting of annotation procedures because embryo morphology assessment can vary across observers, and careful interpretation alongside prior embryo AI work on embryo/oocyte classification, static-image viability prediction, and time-lapse blastocyst prediction. Ethics approval, data privacy safeguards, calibration, and runtime feasibility should also be reported before routine clinical use can be considered.

For future work, we may consider including external out-sample testing, prospective validation, and center-wise performance reporting to better evaluate the generalizability of EMBRACE across imaging systems, laboratory protocols, and clinical workflows. Broader validation is important because prior embryo AI studies have used different task settings, including embryo/oocyte classification, static-image embryo viability prediction, and time-lapse blastocyst prediction. Annotation procedures should also be reported carefully because embryo morphology assessment and time-lapse annotation can vary across observers. Because the current study used a retrospective two-center dataset, broader validation is required before the framework can be considered ready for clinical deployment or routine embryology decision support.

Bootstrap confidence intervals, repeated-run analysis, or other uncertainty estimates should be added for key segmentation and classification metrics to better characterize performance stability. This is particularly important because class imbalance can affect minority-class performance estimates, and sparse fragmentation foregrounds can make segmentation evaluation more sensitive to foreground–background imbalance. Calibration analysis and uncertainty estimation may also help determine whether predicted probabilities are reliable enough for review workflows, particularly for borderline symmetry classes and ambiguous fragmentation cases.

Segmentation improvements should focus on small, low-contrast, and boundary-ambiguous fragments because fragmentation localization remained the major residual-error source in this study. This direction is consistent with the biological difficulty of distinguishing cytoplasmic fragments from surrounding embryo texture. Pixel-level segmentation can support more inspectable fragmentation localization, but ambiguous boundaries remain technically challenging in segmentation, especially when foreground–background imbalance is severe.

Future models may also benefit from integrating static-image morphology with time-lapse morphokinetic data. Time-lapse annotation provides standardized developmental landmarks such as t2 and t4. Early cleavage timing has been associated with blastocyst formation and embryo quality, while morphokinetic parameters have also been linked with implantation-related outcomes. Such extensions should be evaluated carefully using validated data and should remain framed as decision-support research unless prospective clinical outcome validation is performed. Prior embryo AI studies support the feasibility of embryo/oocyte classification, static-image embryo viability prediction, and time-lapse blastocyst prediction, but they do not remove the need for prospective validation before clinical deployment.

## Declarations

### Ethics approval and consent to participate

Two datasets used in this study were collected from two medical centers: Taichung Veterans General Hospital (TCVGH IRB:SE24039A) and Chung Shan Medical University Hospital (CSMUH IRB:CS2-24136), retrospectively.

### Funding

This research did not receive any specific grant from funding agencies in the public, commercial, or not-for-profit sectors.

### Declaration of competing interest

The authors declare that they have no known competing financial interests or personal relationships that could have appeared to influence the work reported in this paper.

### Data availability statement

The private and de-identified EmbryoScope™ dataset used in this study was collected from the Department of Obstetrics and Gynecology at Taichung Veterans General Hospital (VGHTC) between 2021 and 2023. The EmbryoScope™ dataset contains TLI image data and cleavage timing data.

### Code availability statement

Code **availability will be provided upon reasonable request, subject to institutional and ethical restrictions.**

### CRediT authorship contribution statement

Anwar Hussain Sofi: Conceptualization, Methodology, Software, Validation, Formal analysis, Investigation, Data curation, Writing - original draft, Visualization. J.-H. Wang: Supervision, Methodology, Writing - review & editing, Project administration. Ming-Jer Chen: Resources, Clinical supervision, Data curation, Writing - review & editing. Tsung-Hsien Lee: Resources, Clinical supervision, Data curation, Writing - review & editing. Yu-Chiao Yi: Resources, Clinical supervision, Data curation, Writing - review & editing.

### Acknowledgments

Two datasets used in this study were collected from two medical centers: Taichung Veterans General Hospital (TCVGH IRB:SE24039A) and Chung Shan Medical University Hospital (CSMUH IRB:CS2-24136), retrospectively.

## Declaration of generative AI and AI-assisted technologies in manuscript preparation

During the preparation of this work, the authors used ChatGPT to rephrase some sentences. After using this tool, the authors reviewed and edited the content and take full responsibility for the content of the publication.

## References

[1] Ebner T, Moser M, Sommergruber M, Tews G. Selection based on morphological assessment of oocytes and embryos at different stages of preimplantation development: a review. *Hum Reprod Update*. 2003;9(3):251–262. https://doi.org/10.1093/humupd/dmg021.

[2] Alpha Scientists in Reproductive Medicine; ESHRE Special Interest Group of Embryology. The Istanbul consensus workshop on embryo assessment: proceedings of an expert meeting. *Hum Reprod*. 2011;26(6):1270–1283. https://doi.org/10.1093/humrep/der037.

[3] Alikani M, Cohen J, Tomkin G, Garrisi GJ, Mack C, Scott RT. Human embryo fragmentation in vitro and its implications for pregnancy and implantation. *Fertil Steril*. 1999;71(5):836–842. https://doi.org/10.1016/S0015-0282(99)00092-8.

[4] Hardarson T, Hanson C, Sjögren A, Lundin K. Human embryos with unevenly sized blastomeres have lower pregnancy and implantation rates: indications for aneuploidy and multinucleation. *Hum Reprod*. 2001;16(2):313–318. https://doi.org/10.1093/humrep/16.2.313.

[5] Tain DCX, Lim MSR, Ng BL, Hammond E, Wong PS. The influence of Day 2 blastomere symmetry on blastocyst grade and ploidy status. *Fertil Reprod*. 2019;1(2):115–118. https://doi.org/10.1142/S2661318219500117.

[6] Cruz M, Garrido N, Herrero J, Pérez-Cano I, Muñoz M, Meseguer M. Timing of cell division in human cleavage-stage embryos is linked with blastocyst formation and quality. *Reprod Biomed Online*. 2012;25(4):371–381. https://doi.org/10.1016/j.rbmo.2012.06.017.

[7] Alikani M, Calderon G, Tomkin G, Garrisi J, Kokot M, Cohen J. Cleavage anomalies in early human embryos and survival after prolonged culture in vitro. *Hum Reprod*. 2000;15(12):2634–2643. https://doi.org/10.1093/humrep/15.12.2634.

[8] Ciray HN, Campbell A, Agerholm IE, Aguilar J, Chamayou S, Esbert M, et al. Proposed guidelines on the nomenclature and annotation of dynamic human embryo monitoring by a time-lapse user group. *Hum Reprod*. 2014;29(12):2650–2660. https://doi.org/10.1093/humrep/deu278.

[9] Paternot G, Devroe J, Debrock S, D'Hooghe TM, Spiessens C. Intra- and interobserver analysis in the morphological assessment of early-stage embryos. *Reprod Biomed Online*. 2009;19(6):836–844. https://doi.org/10.1186/1477-7827-7-105

[10] Sundvall L, Ingerslev HJ, Breth Knudsen U, Kirkegaard K. Inter- and intra-observer variability of time-lapse annotations. *Hum Reprod*. 2013;28(12):3215–3221. https://doi.org/10.1093/humrep/det366.

[11] Manna C, Nanni L, Lumini A, Pappalardo S. Artificial intelligence techniques for embryo and oocyte classification. *Reprod Biomed Online*. 2013;26(1):42–49. https://doi.org/10.1016/j.rbmo.2012.09.015.

[12] VerMilyea M, Hall JMM, Diakiw SM, Johnston A, Nguyen T, Perugini D, et al. Development of an artificial intelligence-based assessment model for prediction of embryo viability using static images captured by optical light microscopy during IVF. *Hum Reprod*. 2020;35(4):770–784. https://doi.org/10.1093/humrep/deaa013.

[13] Liao Q, Zhang Q, Feng X, et al. Development of deep learning algorithms for predicting blastocyst formation and quality by time-lapse monitoring. *Commun Biol*. 2021;4:415. https://doi.org/10.1038/s42003-021-01937-1.

[14] Yang HY, Leahy BD, Jang WD, Wei D, Kalma Y, Rahav R, et al. BlastAssist: a deep learning pipeline to measure interpretable features of human embryos. *Hum Reprod*. 2024;39(4):698–708. https://doi.org/10.1093/humrep/deae024.

[15] Ronneberger O, Fischer P, Brox T. U-Net: convolutional networks for biomedical image segmentation. *Medical Image Computing and Computer-Assisted Intervention – MICCAI 2015*. LNCS 9351. 2015:234–241. https://doi.org/10.1007/978-3-319-24574-4_28.

[16] Milletari F, Navab N, Ahmadi SA. V-Net: fully convolutional neural networks for volumetric medical image segmentation. *2016 Fourth International Conference on 3D Vision*. 2016:565–571. https://doi.org/10.1109/3DV.2016.79.

[17] Berman M, Triki AR, Blaschko MB. The Lovász-Softmax loss: a tractable surrogate for the optimization of the intersection-over-union measure in neural networks. *Proceedings of the IEEE Conference on Computer Vision and Pattern Recognition*. 2018:4413–4421. https://doi.org/10.1109/CVPR.2018.00464.

[18] Kervadec H, Bouchtiba J, Desrosiers C, Granger E, Dolz J, Ayed IB. Boundary loss for highly unbalanced segmentation. *Med Image Anal*. 2021;67:101851. https://doi.org/10.1016/j.media.2020.101851.

[19] Caruana R. Multitask learning. *Mach Learn*. 1997;28:41–75. https://doi.org/10.1023/A:1007379606734.

[20] Standley T, Zamir A, Chen D, Guibas L, Malik J, Savarese S. Which tasks should be learned together in multi-task learning? *Proceedings of the 37th International Conference on Machine Learning*. PMLR. 2020;119:9120–9132.

[21] Yu T, Kumar S, Gupta A, Levine S, Hausman K, Finn C. Gradient surgery for multi-task learning. *Advances in Neural Information Processing Systems*. 2020;33.

[22] Wang T, Chen Y, Chen J, Liu J, Wu P, Chang S, Lee Y, Su K and Chen C. Diabetic Macular Edema Detection Using End-to-End Deep Fusion Model and Anatomical Landmark Visualization on an Edge Computing Device. *Front. Med.* 2022; 9:851644. doi: 10.3389/fmed.2022.851644

[23] He K, Zhang X, Ren S, Sun J. Deep residual learning for image recognition. *Proceedings of the IEEE Conference on Computer Vision and Pattern Recognition*. 2016:770–778. https://doi.org/10.1109/CVPR.2016.90.

[24] Lin TY, Dollár P, Girshick R, He K, Hariharan B, Belongie S. Feature pyramid networks for object detection. *Proceedings of the IEEE Conference on Computer Vision and Pattern Recognition*. 2017:936–944. https://doi.org/10.1109/CVPR.2017.106.

[25] He H, Garcia EA. Learning from imbalanced data. *IEEE Trans Knowl Data Eng*. 2009;21(9):1263–1284. https://doi.org/10.1109/TKDE.2008.239.

[26] Lin TY, Goyal P, Girshick R, He K, Dollár P. Focal loss for dense object detection. *Proceedings of the IEEE International Conference on Computer Vision*. 2017:2980–2988. https://doi.org/10.1109/ICCV.2017.324.

[27] Pedregosa F, Varoquaux G, Gramfort A, Michel V, Thirion B, Grisel O, et al. Scikit-learn: machine learning in Python. *J Mach Learn Res*. 2011;12:2825–2830.

[28] Meseguer M, Herrero J, Tejera A, Hilligsøe KM, Ramsing NB, Remohí J. The use of morphokinetics as a predictor of embryo implantation. *Hum Reprod*. 2011;26(10):2658–2671. https://doi.org/10.1093/humrep/der256.

[29] Kirkegaard K, Kesmodel U, Hindkjær J, Ingerslev H. Time-lapse parameters as predictors of blastocyst development and pregnancy outcome in embryos from good prognosis patients: a prospective cohort study. *Hum Reprod*. 2013;28(10):2643–2651. https://doi.org/10.1093/humrep/det300.

[30] Bromley J, Guyon I, LeCun Y, Säckinger E, Shah R. Signature verification using a Siamese time delay neural network. *Advances in Neural Information Processing Systems*. 1993;6.

[31] Misra I, Shrivastava A, Gupta A, Hebert M. Cross-stitch networks for multi-task learning. *Proceedings of the IEEE Conference on Computer Vision and Pattern Recognition*. 2016:3994–4003. https://doi.org/10.1109/CVPR.2016.433.

[32] Woo S, Park J, Lee JY, Kweon IS. CBAM: Convolutional Block Attention Module. *European Conference on Computer Vision*. 2018:3–19. https://doi.org/10.1007/978-3-030-01234-2_1.

[33] Cao W, Mirjalili V, Raschka S. Rank consistent ordinal regression for neural networks with application to age estimation. *Pattern Recognit Lett*. 2020;140:325–331. https://doi.org/10.1016/j.patrec.2020.11.008.

[34] Ganin Y, Ustinova E, Ajakan H, Germain P, Larochelle H, Laviolette F, Marchand M, Lempitsky V. Domain-adversarial training of neural networks. *J Mach Learn Res*. 2016;17(59):1–35.

[35] Chen LC, Papandreou G, Schroff F, Adam H. Rethinking atrous convolution for semantic image segmentation. arXiv:1706.05587. 2017. https://doi.org/10.48550/arXiv.1706.05587.

[36] Lee CY, Xie S, Gallagher P, Zhang Z, Tu Z. Deeply-supervised nets. *Proceedings of the 18th International Conference on Artificial Intelligence and Statistics*. PMLR. 2015;38:562–570.

[37] Misera L, Müller-Franzes G, Truhn D, Kather JN. Weakly supervised deep learning in radiology. *Radiology*. 2024;312(1):e232085. https://doi.org/10.1148/radiol.232085.

[38] Wang TY, Deng M, Lee Y, LIU J. Medical image analysis method and device. US Patent 11610306, 2023.